\documentclass[runningheads]{llncs}

\usepackage{eccv}

\usepackage{eccvabbrv}

\usepackage{booktabs}
\usepackage{float}

\usepackage{amsmath,amssymb,amsfonts}
\usepackage{graphicx}
\usepackage{textcomp}
\usepackage{xcolor}
\usepackage{algpseudocode}
\usepackage[english]{babel}
\usepackage[ruled,linesnumbered]{algorithm2e}
\usepackage{caption}

\usepackage[accsupp]{axessibility}  % Improves PDF readability for those with disabilities.

\usepackage{hyperref}

\usepackage{orcidlink}

\begin{document}

\title{Tracking Virtual Meetings in the Wild: Re-identification in Multi-Participant Virtual Meetings} 

\titlerunning{Tracking Virtual Meetings in the Wild}

\author{Oriel Perl\orcidlink{0009-0002-0707-9877} \and
Ido Leshem\orcidlink{0000-0001-7535-4709} \and 
Uria Franko\orcidlink{0009-0008-2196-864X} \and
Yuval Goldman\orcidlink{0009-0002-1060-0774}}

\authorrunning{O.~Perl et al.}

% % TODO FINAL: Replace with your institution list.
\institute{
Novacy\\
\email{\{oriel.net,leshem.ido,uriafranko,goldman1yuval\}@gmail.com}\\
\url{http://www.novacy.io}
}

\maketitle

\begin{abstract}
  In recent years, workplaces and educational institutes have widely adopted virtual meeting platforms. This has led to a growing interest in analyzing and extracting insights from these meetings, which requires effective detection and tracking of unique individuals. In practice, there is no standardization in video meetings recording layout, and how they are captured across the different platforms and services. This, in turn, creates a challenge in acquiring this data stream and analyzing it in a uniform fashion. Our approach provides a solution to the most general form of video recording, usually consisting of a grid of participants  (\cref{fig:videomeeting}) from a single video source with no metadata on participant locations, while using the least amount of constraints and assumptions as to how the data was acquired. Conventional approaches often use YOLO models coupled with tracking algorithms, assuming linear motion trajectories akin to that observed in CCTV footage. However, such assumptions fall short in virtual meetings, where participant video feed window can abruptly change location across the grid. In an organic video meeting setting, participants frequently join and leave, leading to sudden, non-linear movements on the video grid. This disrupts optical flow-based tracking methods that depend on linear motion. Consequently, standard object detection and tracking methods might mistakenly assign multiple participants to the same tracker. In this paper, we introduce a novel approach to track and re-identify participants in remote video meetings, by utilizing the spatio-temporal priors arising from the data in our domain. This, in turn, increases tracking capabilities compared to the use of general object tracking. Our approach reduces the error rate by  \(\sim \)95\%  on average compared to YOLO-based tracking methods as a baseline.

Demo video - \href{https://www.youtube.com/watch?v=Sd9mPqWPZBA} {https://youtu.be/Sd9mPqWPZBA} 

\begin{figure}[h]
\centering
\includegraphics[width=0.4\textwidth]{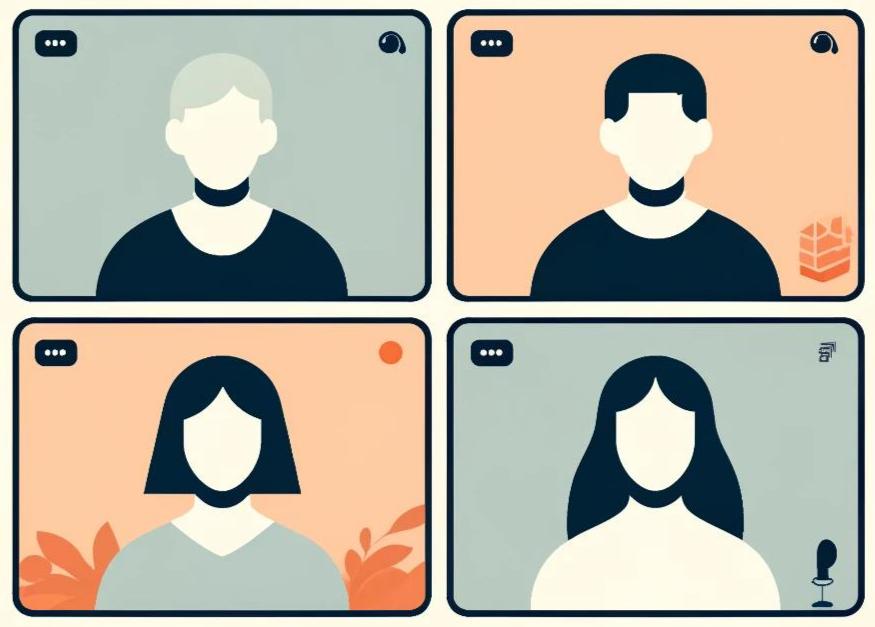}
\caption{Gallery View Illustration - a standard layout for remote meetings, displaying all participants at once.}
\label{fig:videomeeting}
\end{figure}

\footnotetext[1]{Accepted to ECCV 2024 Workshop.}

\end{abstract}

\section{Introduction}

\subsection{Virtual Meetings In Recent Years}
The COVID-19 pandemic has marked the transition towards embracing virtual meetings as a main form of communication and collaboration across industries and verticals. What started as a temporary alternative for face-to-face meetings and classes, rapidly became a fundamental part of our day-to-day communication \cite{hacker2020virtually}. Whether it’s hybrid work, remote classes, or a way for people to connect with each other, remote meeting platforms are here to stay. 

The adoption of virtual meeting platforms such as Zoom, Google Meet, and Microsoft Teams has promoted the incentive for meeting recordings analysis in order to provide valuable insights. For example in the virtual classroom scenario, alerting students’ engagement levels can be highly useful, and in the work-related environment, a more comprehensive analysis could provide valuable insights by tapping into non-verbal cues such as emotional state and the body language displayed by the participants. All of the above require accurate tracking capabilities.

\subsection{Object Tracking Challenges}
The object tracking task aims to provide unique identification for each detected object throughout a series of frames 
\cite{szeliski2022computer}. A successful object tracker would provide a single and consistent ID for each unique object. Object tracking success depends on many factors, among them - image quality, the amount and size of objects, the extent to which their movement is predictable, and the extent to which different objects are occluding each other \cite{yilmaz2006object}. 

\subsection{Challenges In Tracking Participants In Video Meetings}
In order to provide a general solution that is compatible with most recording methods, we addressed the generalized form of video meeting recordings known as “Gallery View” which is comprised of a grid of participants as depicted in \cref{fig:videomeeting}. Performing object tracking in virtual meetings presents the following challenges:
\begin{itemize}
    \item The number of visible participants is dynamic within the video meeting, participants turn their cameras on and off, join and leave the meeting, and move in and out of the frame, introducing various challenges for classic tracking methods.
    \item The number of video feeds does not match the number of participants. i.e two participants can join from a single computer.
    \item Participants can join meetings from different devices, personal computers, mobile phones, or VC room cameras, and these are interchangeable within a meeting, such that participants can start a meeting on one device, and change devices along the call.
    \item The number of participants and length of appearance are dynamic between different video meetings, which limits the ability to select well-generalizing hyper-parameters to be used by clustering methods.
\end{itemize}

These factors lead to a complex video scenario where participants' movements are unpredictable, both spatially within the grid and temporally in their appearances. Consequently, current tracking algorithms often struggle to maintain accuracy and frequently misidentify multiple participants as a single individual or, conversely, incorrectly assign multiple IDs to a single person. Our algorithm delivers an innovative solution designed to preserve the organic behavior of users in video calls while enhancing tracking accuracy. By effectively managing the dynamic nature of video meetings, our approach ensures reliable participant identification, even in complex scenarios involving fluctuating numbers of participants and varied device usage.
\section{Related Work}

% ---------

\subsection{Object-Detection}
Recent advancements in object detection and recognition have significantly improved the accuracy of people tracking across diverse settings \cite{redmon2016you}. The convolutional neural network (CNN) architecture, specifically, has demonstrated superior performance in these tasks, surpassing more traditional approaches \cite{krizhevsky2012imagenet, zhou2019omni}. Among the methodologies employed, YOLO \cite{redmon2016you} stands out for its efficiency and effectiveness in object detection. Additionally, the Multi-Task Cascaded Convolutional Neural Networks (MTCNN) \cite{zhang2016joint} algorithm has gained prominence for its specialized application in face detection.

% ---------

\subsection{Object-Tracking}

\textbf{Classical Approaches}
Traditional object tracking techniques have prominently featured optical flow methods, which calculate motion between two frames at the pixel level, based on the apparent motion of brightness patterns \cite{horn1981determining}. The Lucas-Kanade method utilizes local neighborhood constancy in the flow field to enhance tracking accuracy \cite{lucas1981iterative}. Beyond optical flow, the Kalman Filter \cite{kalman1960new} and the Mean Shift algorithm \cite{comaniciu2002mean} represent significant developments in this area. The Kalman Filter predicts an object’s position using a model of its dynamics, providing a probabilistic approach to motion tracking \cite{kalman1960new}, while the Mean Shift algorithm focuses on finding the densest regions of data points in the feature space, ideal for following objects through complex trajectories \cite{comaniciu2002mean}.

\textbf{Modern Approaches}
Deep learning has revolutionized object tracking, addressing challenges in classical methods through CNNs. Early advances like R-CNN\cite{ren2015faster} improved simultaneous object detection and tracking, while Siamese networks\cite{bertinetto2016fully} enhanced real-time tracking with learned similarity functions. The progression of Multi-Object Tracking (MOT) saw significant milestones: DeepSORT\cite{wojke2017simple} extended the simple online and real-time tracking (SORT)\cite{bewley2016simple} algorithm by incorporating appearance information through a deep association metric. OSNet\cite{zhou2019omni} further advanced re-identification capabilities in complex scenes. Recent algorithms have pushed performance boundaries: ByteTrack\cite{zhang2022bytetrack} improved tracking by utilizing all detection results, not just high-confidence ones; BoTSORT\cite{aharon2022bot} integrated both motion and appearance cues for robust tracking; and StrongSORT\cite{du2023strongsort} further refined these approaches, achieving a 63.5\% HOTA score on the MOT17\cite{dendorfer2021motchallenge} benchmark. These advanced trackers typically combine motion prediction using Kalman Filter\cite{kalman1960new} with deep learning-based re-identification, enabling robust tracking even when objects temporarily disappear from view.

% ---------

\subsection{Advancements In Face Recognition: Detection And Embedding Techniques}

\textbf{Face Detection}
Face detection models have evolved to provide more accurate face localization. RetinaFace\cite{deng2019retinaface} is a single-stage detector that jointly learns face detection, landmark localization, and 3D face reconstruction, proving effective in handling faces at various scales and under challenging conditions. DSFD (Dual Shot Face Detector)\cite{li2019dsfd}  introduces feature enhancement and progressive anchor loss for improved detection across various scales, particularly for small and hard-to-detect faces. MTCNN \cite{zhang2016joint}, while slightly older, remains relevant due to its efficient cascaded architecture that performs face detection, landmark localization, and alignment in a multi-stage process.

\textbf{Face Embedding}
The evolution of face-embedding models has dramatically improved face recognition systems. DeepFace\cite{taigman2014deepface}, an early deep learning approach, significantly narrowed the gap between human and machine performance in face verification. FaceNet\cite{schroff2015facenet} then introduced the concept of mapping facial features into a compact 128D embedding space, enhancing re-identification processes in participant tracking. This approach proved valuable for measuring face similarity and unifying multiple tracking IDs. Further advancements came with SphereFace \cite{liu2017sphereface}, which employs angular softmax loss to learn discriminative features on a hypersphere manifold, and ArcFace \cite{deng2019arcface}, which introduced an additive angular margin loss, achieving state-of-the-art performance on benchmark datasets including LFW \cite{huang2008labeled}. These models have collectively improved robustness to variations in pose, illumination, and expression, advancing the field of face recognition.

The integration of these advanced detection and embedding models has led to more robust face recognition systems, enabling reliable re-identification of individuals across different frames or scenes, even under varying conditions.

% ---------

\subsection{Tracking People In The Wild: CCTV, Gait Embeddings, And Appearance Techniques}
The task of tracking individuals, especially in crowded public spaces, employs diverse methodologies that have significantly evolved due to recent research. Techniques such as gait analysis, which captures unique walking patterns\cite{liao2020model}, and appearance-based models, assessing clothing and body shapes, are critical in contexts ranging from crowd monitoring to urban surveillance \cite{chen2018real, ye2021deep}. Recent innovations have led to the development of more generalized frameworks for tracking, exemplified by the Track Anything Model \cite{yang2023track} (TAM, based on the Segment Anything Model - SAM \cite{kirillov2023segment}). These models provide versatile tools capable of adapting to various dynamic environments, thereby enhancing the robustness and accuracy of tracking systems in handling complex scenarios.

% ---------

\subsection{Virtual Meetings Use-Case}
Research in the field of virtual meetings has explored various techniques aimed at monitoring participants. One notable study employed a method based on video feed window extraction for attention tracking \cite{dacayan2022computer}, Another study explores the use of computer vision to monitor participant attention by tracking eye gaze movements \cite{teka2023towards}. Although these methods offer valuable insights, they sometimes encounter challenges in environments where participants join from the same physical location or when dealing with nonlinear movements of participants within the video feeds.\\

\begin{figure}[t]
\centering
\includegraphics[width=\textwidth]{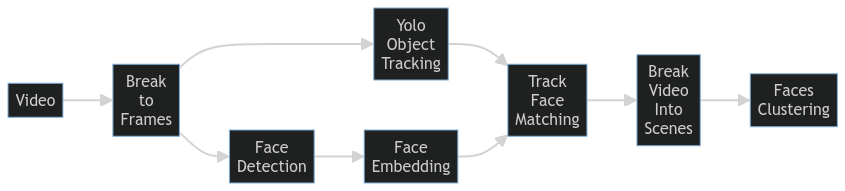}
\caption{Our remote meeting video analysis pipeline: 1) Split video into frames. 2) Run parallel processes for object detection/tracking and face detection/embedding on all frames. 3) Match track IDs with detected faces. 4) Segment video into scenes based on track ID changes. 5) Create latent representations for each track within scenes using embedding vectors. 6) Merge representations across scenes to form full track IDs for meeting participants.}
\label{fig:flowchart}
\end{figure}

To address these challenges, our approach \cref{fig:flowchart} employs advanced open-source tracking tools and face embedding models in conjunction, thus ensuring robust and consistent participant identification throughout meetings. Our solution uniquely identifies individuals, significantly enhancing tracking accuracy and reliability in dynamic virtual meeting scenarios.
\section{Methods}

\subsection{Problem Formulation}
We defined the problem of tracking individuals in virtual meetings as establishing a set of coherent tracks, where each track corresponds to a single participant, and includes the participant's detection in every frame where they appear, such that:

\begin{equation}
F = \{f_0,f_1,...,f_T\} \mid t \in [0,T]
\end{equation}

Where $F$ represents the set of frames extracted from the video, and $T$ is the total number of frames.

Let us denote $O$ as the set of tracks of all participants in the call and $k$ to be the actual number of participants present:

\begin{equation}
 O = \{O_1,O_2,...,O_k\}
 \label{eq:BigO}
\end{equation}

\begin{equation}
 O_i = \{o_{f_t}, o_{f_{t+1}}, \ldots, o_{f_{t+\Delta t}}\}
 \label{eq:O_i}
\end{equation}

 Where $O_i$ is the sequence of detections of participant $i$ in the frame sequence in the different time points $t$ ($t$ can be sparse, as participants can join and leave the call as they please), such that $o_{f_t}$ represents the bounding box of the object $i$ in frame t.

\subsection{Assumptions} 
While analyzing recordings of online meetings introduces some challenges, the nature of an organic meeting allows us to apply some prior assumptions, loosening the constraints on the video analysis:
\begin{itemize}
\item In most types of meetings, participants usually sit close to the camera, this in turn allows us to assume a minimal face size of [40x40] pixels, reducing misdetections.
\item All attendees appear once per frame (from a single camera feed).
\end{itemize}

\subsection{Challenges In Organic Online Meetings}
As mentioned, the task of tracking participants in video meetings holds challenges different from those in traditional tracking tasks. Traditional methods rely on the continuous appearance and the linear trajectory of the detected objects in the video, assumptions that cannot be made in the case of an organic video meeting, where:
\begin{itemize}
    \item Participants can join and leave the meeting and turn their cameras on and off at any given time.
    \item New participants join the meeting unexpectedly.
    \item Every time there is a change in participants or screen sharing, the gallery view grid reshuffles, breaking the continuous motion of the participants.
\end{itemize}

\subsection{Our Solution}
Assuming we are analyzing a video with an unknown number of participants denoted as \( k \), where:
\begin{equation}
P = \{p_1, p_2, \ldots, p_k\} \mid k = \text{number of participants}
\end{equation}
we developed an algorithm to create continuous detected tracks of participants in a virtual meetings scenario. The algorithm follows these steps:

\subsubsection{Breaking Video To Frames}
First, we break the video into frames, for our purposes we chose to use a sampling rate of 2 fps, balancing performance and accuracy:

\begin{equation}
F = \{f_1, f_2, \ldots, f_T\} \mid t \in [0, T]
\end{equation}

Where $F$ represents the set of frames extracted from the video, and $T$ is the total number of frames. For example, in a virtual classroom, alerting students to their engagement levels can be highly beneficial. In a work-related setting, a more comprehensive analysis that considers non-verbal cues, such as emotional states and participants' body language, can provide valuable insights. All of these scenarios rely on accurate tracking capabilities.

\subsubsection{Object Detection And Tracking}
After preprocessing the video into individual frames, we applied object detection and tracking algorithms to identify all the people, establishing a foundation for distinguishing between participants in each frame and generating track IDs. To identify the most suitable approach for this task, our research explored various object detection and tracking algorithms. We reviewed the SSD \cite{liu2016ssd}, RetinaNet \cite{lin2017focal}, and YOLO \cite{redmon2016you} models for object detection. After considering recent surveys \cite{zou2023object}, we chose YOLO for its superior FPS performance (155 fps compared to 6-59 fps for the alternatives) and its simpler implementation and integration with tracking algorithms. For tracking, we combined YOLO with the StrongSORT+OSNet \cite{yolov5-strongsort-osnet-2022} algorithm, as it demonstrated the most consistency in our case.

As mentioned earlier, YOLO with tracking is effective in scenarios like CCTV, where motion is continuous, excluding moments of occlusions. However, in online meetings, participants' video feeds can shift positions abruptly, which presents challenges for the Kalman Filter. This filter assumes linear and continuous motion, potentially leading to track merging errors. Additionally, general bounding box embeddings may struggle to differentiate between participants, especially when facial details are crucial.

Despite these challenges, YOLO-based tracking proved useful for our approach to generate an initial set of participant tracks, where $O$ \cref{eq:BigO} represents all tracks produced, $O_i$ \cref{eq:O_i} is the detected object with track $id_i$, and $o_{f_t}$ denotes the bounding box of object $i$ in frame $t$.

\subsubsection{Face Detection}
After completing the initial tracking, we employed a face detection model to localize the faces within the YOLO-detected bounding boxes. While advanced models like RetinaFace \cite{deng2019retinaface} and DSFD \cite{li2019dsfd} offer superior performance in general scenarios, we chose MTCNN \cite{zhang2016joint} for our pipeline due to several factors that made it an effective choice for our specific scenario:
\begin{itemize}
    \item {\textbf{Simplified use case:} 
    In online meetings, where participants are close to their cameras, MTCNN faces fewer challenges, leading to performance comparable to other methods for our particular needs.}
    \item{\textbf{Streamlined performance and integration:} 
    MTCNN's simple design enabled batch processing for high throughput, while its popularity and high adoption facilitated rapid integration into our pipeline.}
\end{itemize}

To further optimize performance within our framework, we implemented two key heuristics:
\begin{enumerate}
    \item We verified our face detection results against YOLO's superior object detection. We allowed one face per YOLO bounding box, keeping the largest face detection when multiple were found, as active participants likely sit closer to cameras.
    \item Since downstream tasks required a visible face and considering the prior where participants sat relatively close to the camera, we discarded faces smaller than a predefined threshold.
\end{enumerate}
These optimizations matched with our specific use case yielded high-performing results while benefiting from MTCNN's ease of integration and efficiency.

\subsubsection{Face Embedding}

Following our face detection we applied a face-embedding model on each detected face in order to extract 
the participant's distinctive features, this, in turn, allowed us to merge the participant's tracks across the video scenes. We reviewed the leading face-embedding models, with the most common being FaceNet, ArcFace, SphereFace, and DeepFace \cite{schroff2015facenet,deng2019arcface,liu2017sphereface,taigman2014deepface} and their respective performance are denoted in Tab \cref{tab:face_embedding_comparison}. Our scenario involves distinguishing a limited number of participants in meetings, unlike open-world face recognition. This led us to choose FaceNet, which balances performance and efficiency by utilizing 128 dense representations compared to much larger representations in other models. Its proven accuracy (99.63\% on the LFW dataset) and practicality make it ideal for our controlled environment, providing reliable and accurate recognition across a small set of participants.

We generated the sequence of embeddings for each track as denoted:

\begin{equation}
\label{eq:facenet_embed}
Ei = FaceNet(Oi)
\end{equation}

Where $O_i$ \cref{eq:O_i} is the set of the participant $i$ tracks, and we apply FaceNet embedding to each of the detected faces bounding boxes and cluster them within each track.

\begin{table}[t]
\caption{Comparison of Face Embedding Methods on the LFW dataset and their embedding dimensions.}
\centering
\setlength{\tabcolsep}{5pt} % Adjust column separation
\begin{tabular}{c c c c c}
% \hline
\toprule
\textbf{Method} & \textbf{Year} & \textbf{Loss Function} & \textbf{Embedding Size} & \textbf{LFW Accuracy} \\
% \hline
\midrule
DeepFace & 2014 & Softmax & 4096-D & 97.35\% \\
% \hline
FaceNet & 2015 & Triplet Loss & \textbf{128-D} & 99.63\% \\
% \hline
SphereFace & 2017 & Angular Softmax & 512-D & 99.42\% \\
% \hline
ArcFace & 2018 & Additive Angular & 512-D & 99.83\% \\
% \hline
\bottomrule
\end{tabular}

\label{tab:face_embedding_comparison}
\end{table}

\subsubsection{Defining Scenes}
We define a scene as a sequence of frames in which the set of track IDs remains constant. This approach utilizes the provided track IDs to segment the video into distinct scenes. The resulting scenes vary in length: long scenes occur when participants appear continuously in the video without any changes in track IDs, while shorter scenes are produced when changes in the video composition take place, potentially due to participants entering or leaving the frame. In some instances, a track may "break" due to limitations in the initial tracking algorithm's performance.

To meet our specifications, we adopted a high-sensitivity approach towards changes in the tracking log. This decision was based on the rationale that matching similar-faced tracks is a more manageable task than separating a mixed track. By prioritizing sensitivity to track ID changes, we aim to minimize the risk of erroneously combining distinct individuals into a single track, even if this results in occasional over-segmentation of scenes.

\subsubsection{Matching Tracks Across Scenes}
Next, our goal was to match different tracks of faces that appear in the different scenes (a track unique ID consists of a YOLO track ID + scene ID), in order to provide a full sequence of tracked faces throughout the video. Assuming the face embedding of a single person distributes normally in the latent space, the larger the distribution sample, the better the mean of the cluster represents the actual face embedding:
\begin{equation}
Ei\sim Normal(\mu_i,\sigma_i) 
\end{equation}

Therefore, in order to initialize the track matching with an optimal set of embedding representations, we sorted the scenes by the number of frames in descending order while matching between the scenes. This process was done iteratively, such that each new cluster was either assigned to an existing cluster or represented by a new one, using empirical thresholds for optimal performance. This was done to all clusters up to a minimal frames threshold, where the anchoring of the cluster was too noisy to safely assign it to a larger cluster.

\begin{algorithm}[h]
\caption{Online Meeting Participant Tracking Algorithm}
\KwIn{Video}
% \tcc{Break the video into frames}
% \tcc{Process each frame for detection and tracking}
\( F \leftarrow \text{BreakVideoIntoFrames(Video)} \)

\( scenes = \emptyset \)
\( tracks = \emptyset \)
\( objects = \emptyset \)

\For{\( f_i \) in \( F \)}{
    \( updated\_tracks,detected\_objects  \leftarrow \text{YOLO+Strongsort}(f_i) \)
    
    \( objects.append(detected\_objects) \)
    
    \uIf{\( updated\_tracks \neq tracks \)}{
        \( scenes.append(tracks) \)
        
        \( tracks =  updated\_tracks \)
    }
    % \( scenes.append(track\_ids) \) 
}
% \tcc{Detect faces in the bounding boxes identified by YOLO}
\( faces = \emptyset \)

\For{\( o_t \) in \( objects \)}{
    \( detected\_faces \leftarrow \text{MTCNN}(o) \)
    
    \uIf{\( \text{length}(detected\_faces) > 1 \)}{
        \( detected\_faces \leftarrow \text{Select Largest Box}(detected\_faces) \)
    }
    \( faces.append(detected\_faces) \)
}
% \tcc{Encode detected faces}
\( face\_embeddings = \emptyset \)

\ForEach{\( face \) in \( faces \)}{
    \( face\_embedding \leftarrow \text{FaceNet}(face) \)
    
    \( face\_embeddings.append(face\_embedding) \)
}
% \tcc{Sort scenes by the number of frames and match ids across scenes}
\( participants = \emptyset \)

\For{\( scene \) in \( \text{Length\_Sorted}(scenes) \)}{
    \( distance\_matrix \leftarrow distance(scenes_i, scenes_{i+1}) \)
    
    \( min\_distance\_pair = \min(distance\_matrix) \)
    \uIf{\( min\_distance\_pair < threshold \)}{
        \( participants.append(merge(id_1, id_2)) \)
        
        \( distance\_matrix.drop(id\_row, id\_col) \)
    }
}
\KwOut{participants}
\end{algorithm}
\section{Experiments and Results}

\subsection{Dataset}
Our dataset is comprised of a proprietary collection of remote work-related meetings, varying in duration, participant count, and individual presence. Due to the dynamic nature of these meetings, participants frequently joined and left at different times, and may have spontaneously disconnected and reconnected. This behavior affects the spatial arrangement of the "gallery view," which displays all participants simultaneously. The dataset included 30 meetings, with an average length of 25.5 minutes, ranging from 7:18 to 56:26 minutes. The average number of participants per meeting was 5.9, with a minimum of 2 and a maximum of 11 participants, and a standard deviation of 2.32. In total, the dataset comprised 45,901 seconds of virtual meetings, amounting to 91,802 frames analyzed. We denoted the number of unique individuals in a meeting as K whereas in our dataset
$K \in [2, 11]$ and $t_{min} = 120sec$ is the minimum amount of time a participant should be in a meeting for its presence to be counted, otherwise, it will be ignored.

\begin{table}[t]
\caption{Average track IDs Per Number of Participants in the Meeting}
\centering
\setlength{\tabcolsep}{5pt} % Adjust column separation
\resizebox{\textwidth}{!}{
\begin{tabular}{c c c c c c c c c c c}
\toprule
\textbf{Method} & \textbf{Metric} & \textbf{2} & \textbf{3} & \textbf{4} & \textbf{5} & \textbf{6} & \textbf{7} & \textbf{8} & \textbf{9} & \textbf{11} \\
\midrule
YOLOV5+StrongSORT & All tracks & 29.0 & 51.00 & 57.75 & 39.00 & 30.25 & 78.25 & 47.00 & 53.25 & 85.00 \\
YOLOV5+StrongSORT & Track IDs $\geq$ 120s & 5.0 & 9.00 & 10.25 & 6.8 & 8.50 & 11.75 & 13.0 & 23.00 & 20.00 \\
YOLOV8+ByteTrack & All tracks & 17.5 & 21.00 & 21.5 & 14.6 & 22.00 & 22.5 & 28.33 & 39.0 & 59.0 \\
YOLOV8+ByteTrack & Track IDs $\geq$ 120s & 4.5 & 9.67 & 8.25 & 7.0 & 9.5 & 11.0 & 13.67 & 13.50 & 15.00 \\
YOLOV8+BotSort & All tracks & 15.5 & 26.33 & 22.75 & 14.40 & 18.75 & 27.75 & 27.66 & 27.50 & 45.0 \\
YOLOV8+BotSort & Track IDs $\geq$ 120s & 4.5 & 9.66 & 8.25 & 7.0 & 9.5 & 11.0 & 13.66 & 13.5 & 15.0 \\
Our Method & Track IDs $\geq$ 120s & 2.0 & 3.00 & 4.25 & 4.8 & 6.0 & 6.75 & 8.0 & 9.25 & 11.00 \\
\bottomrule
\end{tabular}
}
\label{table:filterd_tracks}
\end{table}

\begin{table}[t]
\caption{MAE for detecting the participant number in a meeting, compared against the ground truth (K), per number of participants in the meeting.}
\centering
\small % Adjust font size here if needed
\setlength{\tabcolsep}{5pt} % Adjust column separation
\begin{tabular}{c c c c c c c c c c}
\toprule
\textbf{Method}  & \textbf{2} & \textbf{3} & \textbf{4} & \textbf{5} & \textbf{6} & \textbf{7} & \textbf{8} & \textbf{9} & \textbf{11} \\
\midrule
YOLOV5 + StrongSORT  & 3.0 & 6.0 & 6.25 & 2.2 & 2.5 & 4.75 & 5.0 & 14.0 & 9.0 \\
YOLOV8 + ByteTrack  & 2.5 & 4.66 & 4.5 & 1.4 & 3.75 & 1.25 & 3.0 & 12.75 & 13.0 \\
YOLOV8 + BotSort  & 2.5 & 6.66 & 4.25 & 2.4 & 3.5 & 4.0 & 5.66 & 4.5 & 4.0 \\
\textbf{Our Method}  & \textbf{0.0} & \textbf{0.0} & \textbf{0.25} & \textbf{0.2} & \textbf{0.0} & \textbf{0.75} & \textbf{0.0} & \textbf{0.25} & \textbf{0.0} \\
\bottomrule
\end{tabular}
\label{table:mae_by_k}
\end{table}

\begin{table}[t]
\caption{Mean Absolute Error (MAE) comparison between different methods for detecting the number of participants in the meeting (K ground truth).}
\centering
\begin{tabular}{c c}
\toprule
\textbf{Method}  & \textbf{MAE} \\
\midrule
YOLOv5 + StrongSORT  & 5.63 \\
YOLOv8 + ByteTrack  & 4.56 \\
YOLOv8 + BotSort  & 4.1 \\
\textbf{Our Method}  & \textbf{0.2} \\
\bottomrule
\end{tabular}
\label{table:methods_mae}
\end{table}

\subsection{Early Experiments}
We initially used a flat approach: applying per-frame face detection and embedding, followed by clustering to form person tracks. While promising on small samples, this method struggled with real-world data due to the varying number of participants and video lengths. This led to difficulties in optimizing clustering parameters, resulting in either under- or over-grouping of individuals.

Our clustering algorithm of choice after some experimentation was HDBSCAN 
\cite{mcinnes2017hdbscan}, a density-based hierarchical clustering method that does not require specifying the number of clusters (K) as it is allocated according to the data. Nonetheless, despite these promising properties, finding the right hyper-parameters that provide consistent results across the diversity of our real-world data proved to be a challenging task, necessitating a new and improved method.

\subsection{Evaluation And Experiment Structure}
Our primary evaluation metric is K, representing the number of unique individuals identified as part of the meeting, which is compared with the ground truth from our labeled data. While this metric may not be sensitive to incorrect ID assignments, wherein some participants are mislabeled as different individuals, it provides a good estimation of the number of unique participants. Ideally, our experiment would also include the Multi-Object Tracking Accuracy (MOTA) metric, which could better reflect our tracking capabilities. However, due to the absence of a benchmark dataset for this specific use case and limited resources for manual tagging, we reported K.

To evaluate the performance of our new method, we compared the difference in K between our method and other common tracking algorithms such as YOLOv5+StrongSORT \cite{yolov5-strongsort-osnet-2022} or YOLOv8+bytesort \cite{Jocher_Ultralytics_YOLO_2023}. The difference in performance between our method and the baseline methods was estimated using the mean absolute error (MAE) measurement. To ensure a reliable comparison, we matched the way we calculate K in both the existing methods and our method. Since we filter out participants who appear for less than 120 seconds, we also filtered out the track IDs that appear for less than this threshold in the existing methods as well. In the results \cref{table:filterd_tracks} we reported both the filtered and unfiltered K values to provide a good estimation of the overall performance.

\begin{figure}[H]
    \centering
    \includegraphics[width=0.95\linewidth]{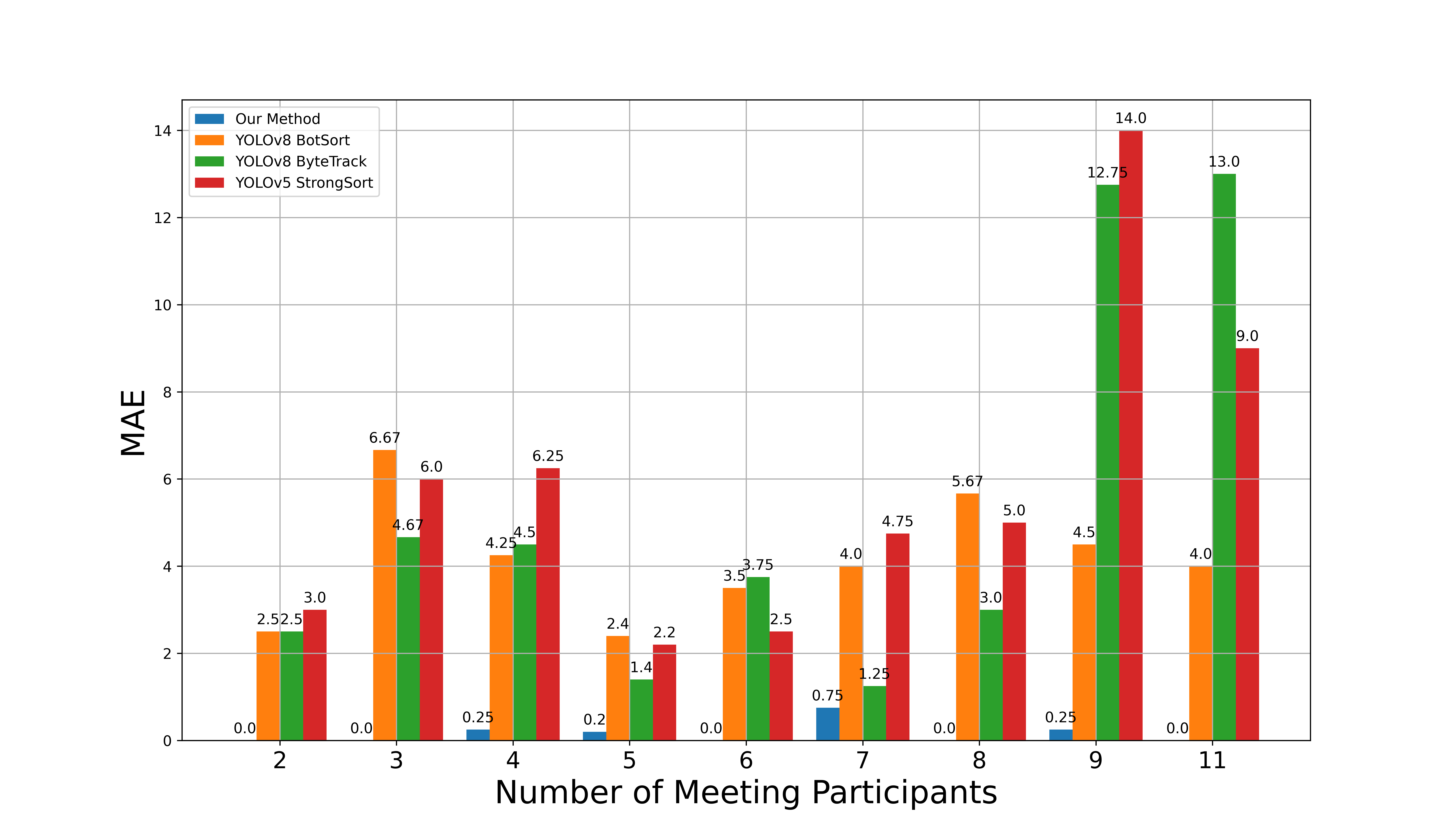}
    \caption{Mean absolute error (MAE) comparing between the predicted number of participants and the actual number of participants (GT)}
    \label{fig:mae_num_of_participants}
\end{figure}

\begin{figure}[H]
    \centering
    \includegraphics[width=0.95\linewidth]{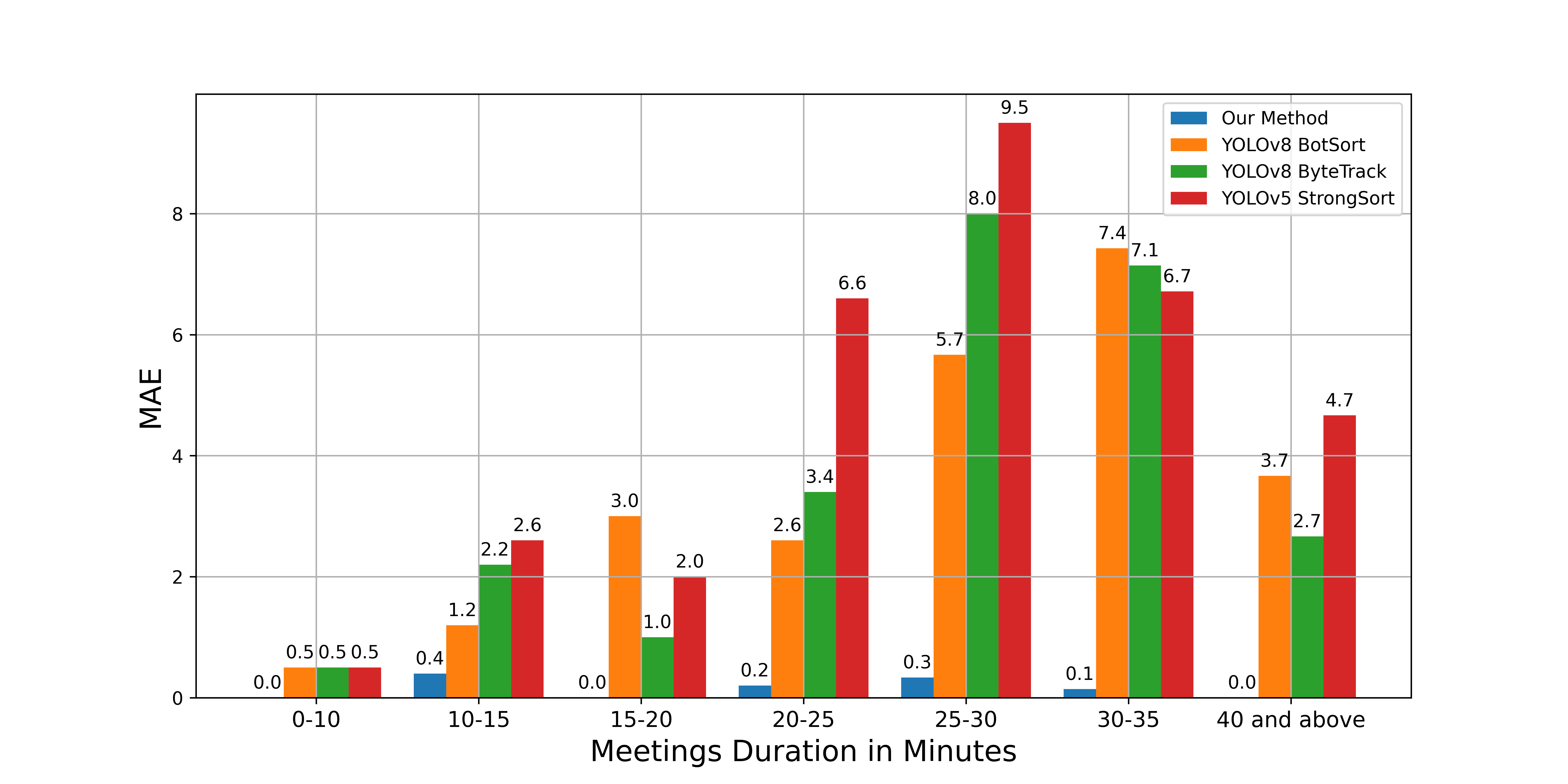}
    \caption{Mean absolute error (MAE) comparing the predicted number of participants and actual number of participants according to meeting duration in minutes.}
    \label{fig:mae_meeting_duration}
\end{figure}

\subsection{Results}

Upon examining the outcomes of our analysis relative to the YOLO-based tracking methods, and taking into account both the number of participants and the duration of the meetings, it is evident that our method exhibits a significant enhancement in performance in both aspects. We conducted a high-level analysis comparing the performance of the methods by calculating the mean absolute error (MAE) across all the data as shown in \cref{table:methods_mae}. It was found that our method had an MAE of 0.2, while the other method's MAE was between 4.1 and 5.63, indicating about  \textbf{\(\sim \)95\% decrease in error rate} compared to baseline.

To get a better understanding of the performance across different numbers of participants, we evaluated the efficacy of the two methods in detecting $K$ by comparing meetings with varying participant counts as shown in \cref{fig:mae_num_of_participants}. For each meeting size, we computed the Mean Absolute Error (MAE) for detecting $K$ using each method. While the other existing methods tended to detect more participants than the ground truth, indicating a tendency towards overestimation, our method demonstrated consistency with the ground truth across all sizes as shown in \cref{table:mae_by_k}.

We then analyzed how the MAE in detecting $K$ varied with the meeting's duration. We divided the meetings into groups based on 5-minute intervals (meetings up to 10 minutes, 10-15 minutes, 15-20 minutes, and so on). 
\cref{fig:mae_meeting_duration} depicts the findings. Finally, We conducted another analysis to count the number of unique track IDs provided by existing methods before and after the 120-second threshold as depicted in \cref{table:filterd_tracks}. This analysis can shed more light on the limitations of the baseline approaches compared to ours.

We believe that the novelty and incremental contribution of our work come from several factors. While our method integrates various components that may not be individually state-of-the-art (SOTA) on their own, the synergistic combination of these elements results in a solution that is both innovative and SOTA. The uniqueness of our approach lies in the way we have orchestrated these components to address the specific challenges of participant tracking in meeting scenarios. Moreover, to the best of our knowledge, our work represents a pioneering effort in addressing this particular problem, as we have not encountered any other methods of tackling these challenges. By introducing this novel approach, we aim to catalyze further research and advancements in this field.

\section{Conclusion}

\subsection{Key Findings}
Our research reveals a critical limitation in existing tracking methods for virtual meeting environments, demonstrating that conventional approaches, including YOLO-based techniques, are inadequate for accurately monitoring online participants. To address this, we developed an innovative method utilizing spatio-temporal priors and face embedding models for tracking and re-identifying participants. This approach significantly outperforms traditional methods, reducing error rates by 95\% and showing increased efficacy as call complexity grows in terms of duration or participant count.
These advancements have far-reaching implications across various domains, including remote work, online education, virtual conferences, etc. Our method provides a foundation for more accurate engagement monitoring, deeper insights into non-verbal communication, and improved meeting analytics. As digital communication continues to evolve, our research not only addresses a critical technical challenge but also paves the way for a new generation of applications that can enhance the effectiveness and understanding of virtual collaborations.

% =================================================

\subsection{Limitations}   
While our approach advances participant tracking for virtual meetings, limitations remain. While our current processing speeds are satisfying for our use case, there is room for improvement to enhance real-time application performance. We believe that with targeted modifications, our approach could be adapted for real-time use. This presents an interesting direction for future research. Challenges with smaller faces in uncontrolled environments such as low-quality video meetings, can affect accuracy, potentially limiting applicability in certain edge-case scenarios. Additionally, due to legal and privacy considerations, we used a proprietary dataset that cannot be shared publicly, and, to the best of our knowledge there are no open-source benchmarks available. We look forward to future advancements in the field and the development of such benchmarks, which would greatly benefit the research community. Addressing these issues and contributing to broader data accessibility remains a priority for future development.

% =================================================

% \clearpage\mbox{}Page \thepage\ of the manuscript.
% \clearpage\mbox{}Page \thepage\ of the manuscript.
% \clearpage\mbox{}Page \thepage\ of the manuscript.
% \clearpage\mbox{}Page \thepage\ of the manuscript.
% \clearpage\mbox{}Page \thepage\ of the manuscript. This is the last page.
% Now we have reached the maximum length of an ECCV \ECCVyear{} submission (excluding references).
% References should start immediately after the main text, but can continue past p.\ 14 if needed.
% \clearpage  % TODO REVIEW/FINAL: This \clearpage needs to be removed from both review and camera-ready versions.

\bibliography{main}

\begin{thebibliography}{10}
\providecommand{\url}[1]{\texttt{#1}}
\providecommand{\urlprefix}{URL }
\providecommand{\doi}[1]{https://doi.org/#1}

\bibitem{aharon2022bot}
Aharon, N., Orfaig, R., Bobrovsky, B.Z.: Bot-sort: Robust associations multi-pedestrian tracking. arXiv preprint arXiv:2206.14651  (2022)

\bibitem{bertinetto2016fully}
Bertinetto, L., Valmadre, J., Henriques, J.F., Vedaldi, A., Torr, P.H.: Fully-convolutional siamese networks for object tracking. In: Computer Vision--ECCV 2016 Workshops: Amsterdam, The Netherlands, October 8-10 and 15-16, 2016, Proceedings, Part II 14. pp. 850--865. Springer (2016)

\bibitem{bewley2016simple}
Bewley, A., Ge, Z., Ott, L., Ramos, F., Upcroft, B.: Simple online and realtime tracking. In: 2016 IEEE international conference on image processing (ICIP). pp. 3464--3468. IEEE (2016)

\bibitem{yolov5-strongsort-osnet-2022}
Broström, M.: Real-time multi-camera multi-object tracker using yolov5 and strongsort with osnet. \url{https://github.com/mikel-brostrom/Yolov5_StrongSORT_OSNet} (2022)

\bibitem{chen2018real}
Chen, L., Ai, H., Zhuang, Z., Shang, C.: Real-time multiple people tracking with deeply learned candidate selection and person re-identification. In: 2018 IEEE international conference on multimedia and expo (ICME). pp.~1--6. IEEE (2018)

\bibitem{comaniciu2002mean}
Comaniciu, D., Meer, P.: Mean shift: A robust approach toward feature space analysis. IEEE Transactions on pattern analysis and machine intelligence  \textbf{24}(5),  603--619 (2002)

\bibitem{dacayan2022computer}
Dacayan, T., Kwak, D., Zhang, X.: Computer-vision based attention monitoring for online meetings. In: 2022 5th International Conference on Pattern Recognition and Artificial Intelligence (PRAI). pp. 533--538. IEEE (2022)

\bibitem{dendorfer2021motchallenge}
Dendorfer, P., Osep, A., Milan, A., Schindler, K., Cremers, D., Reid, I., Roth, S., Leal-Taix{\'e}, L.: Motchallenge: A benchmark for single-camera multiple target tracking. International Journal of Computer Vision  \textbf{129},  845--881 (2021)

\bibitem{deng2019arcface}
Deng, J., Guo, J., Xue, N., Zafeiriou, S.: Arcface: Additive angular margin loss for deep face recognition. In: Proceedings of the IEEE/CVF conference on computer vision and pattern recognition. pp. 4690--4699 (2019)

\bibitem{deng2019retinaface}
Deng, J., Guo, J., Zhou, Y., Yu, J., Kotsia, I., Zafeiriou, S.: Retinaface: Single-stage dense face localisation in the wild. arXiv preprint arXiv:1905.00641  (2019)

\bibitem{du2023strongsort}
Du, Y., Zhao, Z., Song, Y., Zhao, Y., Su, F., Gong, T., Meng, H.: Strongsort: Make deepsort great again. IEEE Transactions on Multimedia  \textbf{25},  8725--8737 (2023)

\bibitem{hacker2020virtually}
Hacker, J., Vom~Brocke, J., Handali, J., Otto, M., Schneider, J.: Virtually in this together--how web-conferencing systems enabled a new virtual togetherness during the covid-19 crisis. European Journal of Information Systems  \textbf{29}(5),  563--584 (2020)

\bibitem{horn1981determining}
Horn, B.K., Schunck, B.G.: Determining optical flow. Artificial intelligence  \textbf{17}(1-3),  185--203 (1981)

\bibitem{huang2008labeled}
Huang, G.B., Mattar, M., Berg, T., Learned-Miller, E.: Labeled faces in the wild: A database forstudying face recognition in unconstrained environments. In: Workshop on faces in'Real-Life'Images: detection, alignment, and recognition (2008)

\bibitem{Jocher_Ultralytics_YOLO_2023}
Jocher, G., Chaurasia, A., Qiu, J.: {Ultralytics YOLO} (Jan 2023), \url{https://github.com/ultralytics/ultralytics}

\bibitem{kalman1960new}
Kalman, R.E.: A new approach to linear filtering and prediction problems  (1960)

\bibitem{kirillov2023segment}
Kirillov, A., Mintun, E., Ravi, N., Mao, H., Rolland, C., Gustafson, L., Xiao, T., Whitehead, S., Berg, A.C., Lo, W.Y., et~al.: Segment anything. In: Proceedings of the IEEE/CVF International Conference on Computer Vision. pp. 4015--4026 (2023)

\bibitem{krizhevsky2012imagenet}
Krizhevsky, A., Sutskever, I., Hinton, G.E.: Imagenet classification with deep convolutional neural networks. Advances in neural information processing systems  \textbf{25} (2012)

\bibitem{li2019dsfd}
Li, J., Wang, Y., Wang, C., Tai, Y., Qian, J., Yang, J., Wang, C., Li, J., Huang, F.: Dsfd: dual shot face detector. In: Proceedings of the IEEE/CVF conference on computer vision and pattern recognition. pp. 5060--5069 (2019)

\bibitem{liao2020model}
Liao, R., Yu, S., An, W., Huang, Y.: A model-based gait recognition method with body pose and human prior knowledge. Pattern Recognition  \textbf{98},  107069 (2020)

\bibitem{lin2017focal}
Lin, T.Y., Goyal, P., Girshick, R., He, K., Doll{\'a}r, P.: Focal loss for dense object detection. In: Proceedings of the IEEE international conference on computer vision. pp. 2980--2988 (2017)

\bibitem{liu2016ssd}
Liu, W., Anguelov, D., Erhan, D., Szegedy, C., Reed, S., Fu, C.Y., Berg, A.C.: Ssd: Single shot multibox detector. In: Computer Vision--ECCV 2016: 14th European Conference, Amsterdam, The Netherlands, October 11--14, 2016, Proceedings, Part I 14. pp. 21--37. Springer (2016)

\bibitem{liu2017sphereface}
Liu, W., Wen, Y., Yu, Z., Li, M., Raj, B., Song, L.: Sphereface: Deep hypersphere embedding for face recognition. In: Proceedings of the IEEE conference on computer vision and pattern recognition. pp. 212--220 (2017)

\bibitem{lucas1981iterative}
Lucas, B.D., Kanade, T.: An iterative image registration technique with an application to stereo vision. In: IJCAI'81: 7th international joint conference on Artificial intelligence. vol.~2, pp. 674--679 (1981)

\bibitem{mcinnes2017hdbscan}
McInnes, L., Healy, J., Astels, S., et~al.: hdbscan: Hierarchical density based clustering. J. Open Source Softw.  \textbf{2}(11), ~205 (2017)

\bibitem{redmon2016you}
Redmon, J., Divvala, S., Girshick, R., Farhadi, A.: You only look once: Unified, real-time object detection. In: Proceedings of the IEEE conference on computer vision and pattern recognition. pp. 779--788 (2016)

\bibitem{ren2015faster}
Ren, S., He, K., Girshick, R., Sun, J.: Faster r-cnn: Towards real-time object detection with region proposal networks. Advances in neural information processing systems  \textbf{28} (2015)

\bibitem{schroff2015facenet}
Schroff, F., Kalenichenko, D., Philbin, J.: Facenet: A unified embedding for face recognition and clustering. In: Proceedings of the IEEE conference on computer vision and pattern recognition. pp. 815--823 (2015)

\bibitem{szeliski2022computer}
Szeliski, R.: Computer vision: algorithms and applications. Springer Nature (2022)

\bibitem{taigman2014deepface}
Taigman, Y., Yang, M., Ranzato, M., Wolf, L.: Deepface: Closing the gap to human-level performance in face verification. In: Proceedings of the IEEE conference on computer vision and pattern recognition. pp. 1701--1708 (2014)

\bibitem{teka2023towards}
Teka, K., Shastri, D.: Towards automatic detection of participant attention in virtual meetings. In: 2023 Congress in Computer Science, Computer Engineering, \& Applied Computing (CSCE). pp. 2731--2733. IEEE (2023)

\bibitem{wojke2017simple}
Wojke, N., Bewley, A., Paulus, D.: Simple online and realtime tracking with a deep association metric. In: 2017 IEEE international conference on image processing (ICIP). pp. 3645--3649. IEEE (2017)

\bibitem{yang2023track}
Yang, J., Gao, M., Li, Z., Gao, S., Wang, F., Zheng, F.: Track anything: Segment anything meets videos. arXiv preprint arXiv:2304.11968  (2023)

\bibitem{ye2021deep}
Ye, M., Shen, J., Lin, G., Xiang, T., Shao, L., Hoi, S.C.: Deep learning for person re-identification: A survey and outlook. IEEE transactions on pattern analysis and machine intelligence  \textbf{44}(6),  2872--2893 (2021)

\bibitem{yilmaz2006object}
Yilmaz, A., Javed, O., Shah, M.: Object tracking: A survey. Acm computing surveys (CSUR)  \textbf{38}(4),  13--es (2006)

\bibitem{zhang2016joint}
Zhang, K., Zhang, Z., Li, Z., Qiao, Y.: Joint face detection and alignment using multitask cascaded convolutional networks. IEEE signal processing letters  \textbf{23}(10),  1499--1503 (2016)

\bibitem{zhang2022bytetrack}
Zhang, Y., Sun, P., Jiang, Y., Yu, D., Weng, F., Yuan, Z., Luo, P., Liu, W., Wang, X.: Bytetrack: Multi-object tracking by associating every detection box. In: European conference on computer vision. pp. 1--21. Springer (2022)

\bibitem{zhou2019omni}
Zhou, K., Yang, Y., Cavallaro, A., Xiang, T.: Omni-scale feature learning for person re-identification. In: Proceedings of the IEEE/CVF international conference on computer vision. pp. 3702--3712 (2019)

\bibitem{zou2023object}
Zou, Z., Chen, K., Shi, Z., Guo, Y., Ye, J.: Object detection in 20 years: A survey. Proceedings of the IEEE  \textbf{111}(3),  257--276 (2023)

\end{thebibliography}
\bibliographystyle{splncs04}
\end{document}